\documentclass{article}
\usepackage{mystyle}
\usepackage[utf8]{inputenc}
\usepackage{soul}
\usepackage{mathptmx}
\usepackage[scaled=0.9]{helvet}
\usepackage{courier}
\usepackage[T1]{fontenc}

\usepackage{url} 
\usepackage[small]{caption}
\usepackage{graphicx}
\usepackage{amsmath}
\usepackage{booktabs}

\usepackage{amssymb}
\usepackage{xspace}
\usepackage{mathtools}
\usepackage{microtype}
\usepackage[inline]{enumitem}
\usepackage{nicefrac}
\usepackage{tikz}
\usepackage{pgfplots}
\usepackage[hidelinks]{hyperref}

\pgfplotsset{compat=1.16}
\usepgfplotslibrary{groupplots}

\newcommand{\nsal}{\textsc{NeSAL}\xspace}
\newcommand{\verifier}{\textsc{NSV}\xspace}

\newcommand{\pre}{\ensuremath{\varphi_\mathit{pre}}}
\newcommand{\assign}{\ensuremath{\varphi_\mathit{assign}}}
\newcommand{\post}{\ensuremath{\varphi_\mathit{post}}}

\DeclareMathOperator*{\argmax}{arg\,max}

\newtheorem{theorem}{Theorem}
\newtheorem{lemma}{Lemma}

\hypersetup{
	pdftitle = {Neuro-Symbolic Verification of Deep Neural Networks},
	pdfauthor = {Xuan Xie, Kristian Kersting, Daniel Neider},
	pdfkeywords = {Verification, Deep Neural Networks, Neuro-Symbolic Reasoning},
}

\title{Neuro-Symbolic Verification of Deep Neural Networks}

\author{
	Xuan Xie\textsuperscript{1} \and
	Kristian Kersting\textsuperscript{2} \And
	Daniel Neider\textsuperscript{1}\footnote{Contact author (\href{mailto:neider@mpi-sws.org}{neider@mpi-sws.org})} \\
	\affiliations
	\textsuperscript{1}Max Planck Institute for Software Systems, Kaiserslautern Germany \\
	\textsuperscript{2}Department of Computer Science, TU Darmstadt, Darmstadt, Germany \\
}

\begin{document}

%---------- Activate page numbers ----------
\pagestyle{plain}

%---------- Make title ----------
\maketitle

%---------- Abstract ----------
\begin{abstract}
Formal verification has emerged as a powerful approach to ensure the safety and reliability of deep neural networks.
However, current verification tools are limited to only a handful of properties that can be expressed as first-order constraints over the inputs and output of a network. While adversarial robustness and fairness fall under this category, many real-world properties (e.g., ``an autonomous vehicle has to stop in front of a stop sign'') remain outside the scope of existing verification technology.
To mitigate this severe practical restriction, we introduce a novel framework for verifying neural networks, named neuro-symbolic verification.
The key idea is to use neural networks as part of the otherwise logical specification, enabling the verification of a wide variety of complex, real-world properties, including the one above.
Moreover, we demonstrate how neuro-symbolic verification can be implemented on top of existing verification infrastructure for neural networks, making our framework easily accessible to researchers and practitioners alike.
\end{abstract}

%---------- Intro ----------
% !Tex root=main.tex

\section{Introduction}
\label{sec:intro}

The exceptional performance of deep neural networks in areas such as perception and natural language processing has made them an integral part of many real-world AI systems, including safety-critical ones such as medical diagnosis and autonomous driving.
However, neural networks are inherently opaque, and various defects have been found in state-of-the-art networks.
The perhaps best-known one among those is the lack of adversarial robustness~\cite{DBLP:journals/corr/SzegedyZSBEGF13}.
This term describes the phenomenon that even slight perturbations of an input to a neural network can cause entirely different outputs.
In fact, the prevalence of defects in learning-based systems has prompted the introduction of a dedicated database to monitor AI incidents and avoid repeated undesired outcomes~\cite{DBLP:conf/aaai/McGregor21}.%
\footnote{\url{https://incidentdatabase.ai/}}

Motivated by decade-long advances in software reliability, formal verification has emerged as a powerful approach to ensure the correctness and safety of neural networks (we refer the reader to Section~\ref{sec:background} for further details).
In contrast to (empirical) statistical evaluation methods from machine learning, such as  cross-validation, formal verification techniques have the great advantage that they are not limited to checking a given property on just a finite number of inputs.
Instead, they allow verifying that a property holds for all (or at least infinitely many) inputs to a deep neural network, including unseen data and corner cases.
However, formal verification has a fundamental limitation that often constitutes a significant obstacle in practice: it requires that the property to verify can be expressed as ``simple'' (typically quantifier-free, first-order) constraints over the inputs and output of the neural network.

While adversarial robustness and fairness fall under the above category, many real-world properties remain outside the scope of existing verification technology.
Consider, for instance,  a deep neural network controlling an autonomous vehicle and the property that the vehicle needs to decelerate as soon as a stop sign appears in the front view.
It is not hard to see that formalizing this property in terms of constraints on inputs and outputs of a deep neural network is extremely difficult, if not impossible: it would require capturing all essential features of all possible stop signs, such as their position, shape, and orientation, on the level of image pixels.
If this was possible, machine learning would not be necessary in the first place because one could implement a detection algorithm based on such a formal specification.

To overcome this severe limitation and make formal verification applicable to real-world scenarios, we propose a neuro-symbolic framework for verifying deep neural networks.
Following the general idea of neuro-symbolic reasoning~\cite{DBLP:journals/flap/GarcezGLSST19,DBLP:conf/ijcai/RaedtDMM20}, our key contribution is a novel specification language, named \emph{Neuro-Symbolic Assertion Language (\nsal)}, that allows one to combine logical specifications and deep neural networks.
The neural networks, which we call \emph{specification networks}, serve as proxies for complex, semantic properties and enable the integration of advances in fields such as perception and natural language recognition into formal verification.
In the context of our example above, one could train a highly-specialized specification network to detect stop signs.
Then, the desired property can be expressed straightforwardly as ``if the specification network detects a stop sign, the network controlling the autonomous vehicle has to issue a braking command''.
We present our neuro-symbolic framework in Section~\ref{sec:neuro-symbolic-verification}, where we also discuss ways of obtaining specification networks in practice.

An essential feature of our framework is that it can be built on top of the existing verification infrastructure for neural networks, which we demonstrate in Section~\ref{sec:reduction-to-deductive-verification} by presenting a prototype verifier for \nsal properties based on the popular Marabou framework~\cite{DBLP:conf/cav/KatzHIJLLSTWZDK19}.
In Section~\ref{sec:evaluation}, we then show that our prototype effectively verifies a variety of neuro-symbolic properties and can produce informative counterexamples to failed verification attempts.
As targets for our verification, we have trained deep neural networks on the German Traffic Sign Recognition Benchmark (GTSRB)~\cite{DBLP:conf/ijcnn/StallkampSSI11} and MNIST~\cite{lecun2010mnist}.

Finally, we want to highlight that our neuro-symbolic framework is general and not limited to deep neural networks.
Instead, it can---in principle---be applied to any system that allows for suitable verification techniques, including differential models, hardware, and software.
However, we leave an in-depth study of this promising direction to future work.

%---------- Related Work ----------
% !Tex root=main.tex

\subsubsection*{Related Work}
Driven by the demand for trustworthy and reliable artificial intelligence, the formal verification of deep neural networks has become a very active and vibrant research area over the past five years (we refer the reader to a textbook~\cite{albarghouthi-book} for a detailed overview).
To the best of our knowledge, Seshia \textit{et al.}~\shortcite{DBLP:conf/atva/SeshiaDDFGKSVY18} conducted the first comprehensive survey of correctness properties arising in neural network verification.
Similar to the verification of software, the authors classify these properties into several (not necessarily disjoint) categories: system-level specifications,  input-output robustness,  other input-output relations, semantic invariance, monotonicity, fairness, input\slash{}distributional assumptions, coverage criteria, and temporal specifications.
However, we are not aware of any work proposing a neuro-symbolic verification approach, neither for deep neural networks or other differential models nor for hardware or software.

The key motivation of neuro-symbolic AI is to combine the advantages of symbolic and deep neural representations into a joint system~\cite{DBLP:series/cogtech/GarcezLG2009,DBLP:journals/flap/GarcezGLSST19,DBLP:conf/nips/JiangA20,DBLP:journals/corr/abs-2012-05876}.
This is often done in a hybrid fashion where a neural network acts as a perception module that interfaces with a symbolic reasoning system~\cite{DBLP:conf/nips/Yi0G0KT18,DBLP:conf/iclr/MaoGKTW19}.
The goal of such an approach is to mitigate the issues of one type of representation by the other (e.g., using the power of symbolic reasoning to handle the generalizability issues of neural networks and to handle the difficulty of noisy data for symbolic systems via neural networks). 
Recent work has also demonstrated the advantage of neuro-symbolic XAI~\cite{DBLP:conf/ijcai/CiravegnaGGMM20,DBLP:conf/cvpr/StammerSK21} and commonsense reasoning~\cite{DBLP:conf/aaai/ArabshahiLGMAM21}.
The link to verification, however, has not been explored much.
Yang \textit{et al.}~\shortcite{DBLP:journals/fac/YangLLHLCHZ21} explore symbolic propagation, but a higher-order specification framework does not exist. 

To automatically verify correctness properties of deep neural networks, a host of distinct techniques have been proposed.
The arguably most promising and, hence, most popular ones are \emph{abstract interpretation}~\cite{DBLP:conf/sp/GehrMDTCV18,DBLP:conf/nips/SinghGMPV18,DBLP:conf/pldi/BonaertDBV21,DBLP:conf/ijcai/HenriksenL21} and \emph{deductive verification}~\cite{DBLP:conf/atva/Ehlers17,DBLP:conf/cav/KatzHIJLLSTWZDK19}.
The former performs the computation of a network abstractly on an infinite number of inputs, while the latter reduces the verification problem to a satisfiability check of logic formulas (we survey both in Section~\ref{sec:background}).
In addition, various other approaches have been suggested, which are often derived from existing techniques for software verification.
Examples include optimization-based methods~\cite{DBLP:conf/ijcai/GowalDSMK19},
concolic testing~\cite{DBLP:conf/kbse/SunWRHKK18},
model checking~\cite{DBLP:journals/jcst/LiuSZW20},
refinement types~\cite{DBLP:conf/aplas/KokkeKKAA20},
and decomposition-based methods~\cite{DBLP:conf/ijcai/KouvarosL21,DBLP:conf/ijcai/BattenKLZ21}.
While our neuro-symbolic framework is independent of the actual verification technique, we focus on deductive verification in this paper.
As mentioned above, we leave an in-depth study of which other techniques can benefit from our approach for future work.

%---------- Background ----------
% !Tex root=main.tex

\section{Background on Neural Network Verification}
\label{sec:background}

Neural Network Verification is the task of formally proving that a deep neural network satisfies a semantic property (i.e., a property that refers to the semantic function computed by the network).
To not clutter this section with too many technical details, let us illustrate this task through two popular examples: adversarial robustness and fairness.
We will later formalize neural network verification in Section~\ref{sec:neuro-symbolic-verification} when we introduce our neuro-symbolic verification framework.

In the case of adversarial robustness, one wants to prove that a neural network is robust to small perturbations of its inputs (i.e., that small changes to an input do not change the output).
To make this mathematically precise, let us assume that we are given a multi-class neural network $f \colon \mathbb R^m \to \{ c_1, \ldots, c_n \}$ with $m$ features and $n$ classes, a specific input $\vec{x}^\star \in \mathbb R^m$, a distance function $\mathit{d} \colon \mathbb R^m \times \mathbb R^m \to \mathbb R_+$, and a distance $\varepsilon \geq 0$.
Then, the task is to prove that 
\begin{equation}
	\label{eq:adversarial-robustness}
	\mathit{d}(\vec{x}^\star, \vec{x}) \leq \varepsilon \text{ implies } f(\vec{x}^\star) = f(\vec{x}) 
\end{equation}
for all inputs $\vec{x} \in \mathbb R^m$.
In other words, the classes of $\vec{x}^\star$ and every input at most $\varepsilon$ away from $\vec{x}^\star$ must coincide.
An input $\vec{x} \in \mathbb R^m$ violating Property~\eqref{eq:adversarial-robustness} is called an adversarial example and witnesses that $f$ is not adversarially robust.

In the case of fairness, one wants to prove that the output of a neural network is not influenced by a sensitive feature such as sex or race. Again, let us assume that we are given a neural network $f \colon \mathbb R^m \to \mathbb R^n$ with $m$ features, including a sensitive feature $i \in \{ 1, \ldots, m \}$.
Then, the task is to prove that
\begin{equation}
	\label{eq:fairness}
	x_i \neq x'_i \land \bigwedge_{j \neq i} x_j = x'_j \text{ implies } f(\vec{x}) = f(\vec{x}') 
\end{equation}
for all pairs $\vec{x}, \vec{x}' \in \mathbb R^m$ of inputs with $\vec{x} = (x_1, \ldots, x_m)$ and $\vec{x}' = (x'_1, \ldots x'_m)$.
In other words, if two inputs $\vec{x}$ and $\vec{x}'$ only differ on a sensitive feature, then the output of $f$ must not change.
Note that in the case of fairness, a counterexample consists of pairs $\vec{x}, \vec{x}'$ of inputs.

Properties~\eqref{eq:adversarial-robustness} and \eqref{eq:fairness} demonstrate a fundamental challenge of neural network verification: the task is to prove a property for \emph{all} (usually infinitely many) inputs.
Thus, cross-validation or other statistical approaches from machine learning are no longer sufficient because they test the network only on a finite number of inputs. Instead, one needs to employ methods that can reason symbolically about a given network.

Motivated by the success of modern software verification, a host of symbolic methods for the verification of neural networks have been proposed recently~\cite{albarghouthi-book}.
Among the two most popular are deductive verification~\cite{DBLP:conf/atva/Ehlers17,DBLP:conf/cav/KatzHIJLLSTWZDK19} and abstract interpretation~\cite{DBLP:conf/sp/GehrMDTCV18,DBLP:conf/nips/SinghGMPV18}.
Let us briefly sketch both.

The key idea of deductive verification is to compile a deep neural network $f \colon \mathbb R^m \to \mathbb R^n$ together with a semantic property $P$ into a logic formula $\psi_{f, P}$, called \emph{verification condition}.
This formula typically falls into the quantifier-free fragment of real arithmetic and is designed to be valid (i.e., satisfied by all inputs) if and only if $f$ satisfies $P$.
To show the validity of $\psi_{f, P}$, one checks whether its negation $\lnot \psi_{f, P}$ is satisfiable.
This can be done either with the help of an off-the-shelf Satisfiability Modulo Theory solver (such as Z3~\cite{DBLP:conf/tacas/MouraB08}) or using one of the recently proposed, specialized constraint solvers such as Planet~\cite{DBLP:conf/atva/Ehlers17} or Marabou~\cite{DBLP:conf/cav/KatzHIJLLSTWZDK19}.
If $\lnot \psi_{f, P}$ is unsatisfiable, then $\psi_{f, P}$ is valid, and---by construction---$f$ satisfies $P$.
If $\lnot \psi_{f, P}$ is satisfiable, on the other hand, then $\psi_{f, P}$ is not valid, implying that $f$ violates the property $P$.
In the latter case, most constraint solvers (including the ones mentioned above) can produce an assignment satisfying $\lnot \psi_{f, P}$, which can then be used to extract inputs to $f$ that witness a violation of $P$.

Abstract interpretation is a mathematical framework for computing sound and precise approximations of the semantics of software and other complex systems~\cite{DBLP:conf/popl/CousotC77}.
When applied to neural network verification, the basic idea is to over-approximate the computation of a deep neural network on an infinite set of inputs.
Each such infinite set is symbolically represented by an element of a so-called \emph{abstract domain}, which consists of logical formulas capturing shapes such as $n$-dimensional boxes, polytopes, or zonotopes. 
To approximate a network's computation, an element of the abstract domain is propagated through the layers of the network.
Since layers operate on concrete values and not abstract elements, this propagation requires replacing each layer with an abstract one (called abstract transformer) that computes the effects of the layer on abstract elements.
Thus, when given an abstract element $A$ in the input space of a network $f$ (e.g., representing the neighborhood of a fixed input $\vec{x}^\star$), the result of abstract interpretation is an abstract element $A'$ in the output space over-approximating all outputs $f(x)$ of concrete inputs $x \in A$.
To verify that a property $P$ holds, it is then enough to check whether $A'$ is included in an abstract element representing all outputs satisfying $P$.
Since abstract interpretation computes over-approximations of the actual input-output behavior of a network, the property $P$ typically describes a safety condition.

While neural network verification is a vibrant and highly active field, virtually all existing research suffers from three substantial shortcomings:
\begin{enumerate}
	\item
	Existing research focuses on verifying ``simple'' properties that can be formalized using quantifier-free first-order constraints on the inputs and outputs of a network.
	Examples of such properties include adversarial robustness and fairness, illustrated by Properties~\eqref{eq:adversarial-robustness} and \eqref{eq:fairness} above.
	However, the overwhelming majority of relevant correctness properties cannot be expressed in this simple way.
	As an example, consider a neural network controlling an autonomous car and the property that the car needs to decelerate as soon as a stop sign appears in the front view.
	It is clear that formalizing this property is extremely hard (if not impossible, as Seshia and Sadigh~\shortcite{DBLP:journals/corr/SeshiaS16} argue):
	it would require us to mathematically capture all essential features of all possible stop signs, including their position, shape, angle, color, etc.
	\item
	Virtually all properties considered in neural network verification today are either \emph{local} (referring to inputs in the neighborhood of an a~priori fixed input $\vec{x}^\star$) or \emph{global} (referring to all inputs).
	Adversarial robustness is an example of the former type, while fairness illustrates the latter.
	However, a more natural and helpful approach would be to restrict the verification to inputs from the underlying data distribution since we do typically not expect our networks to process out-of-distribution data.
	Again, such a restriction is very hard to capture mathematically and, therefore, not featured by current methods.
	\item
	A fundamental problem, especially when verifying global properties, is that counterexamples (i.e., inputs witnessing the violation of the property) are often out of distribution and, hence, of little value.
	Again, restricting the verification to inputs originating from the underlying data distribution would mitigate this issue but is not supported by current approaches.
\end{enumerate}
In the next section, we address these drawbacks by introducing a neuro-symbolic framework for neural network verification.

%---------- Neuro-Symbolic Verification ----------
% !Tex root=main.tex

\section{A Neuro-Symbolic Verification Framework}
\label{sec:neuro-symbolic-verification}

As illustrated by Properties~\eqref{eq:adversarial-robustness} and \eqref{eq:fairness}, the primary obstacle in today's neural network verification is that correctness properties have to be formalized in a suitable---often relatively simple---logical formalism that relates inputs and outputs of a deep neural network (e.g., the quantifier-free fragment of real arithmetic).
This requirement fundamentally limits current verification approaches to only a few different types of correctness properties, arguably making them ill-equipped to tackle real-world AI verification tasks.

As a first step towards overcoming this severe practical limitation, we propose a neuro-symbolic approach to neural network verification.
Our main idea is seemingly simple yet powerful: we propose the use of highly specialized deep neural networks, named \emph{specification networks}, as proxy objects for capturing semantic correctness properties.
We introduce the concept of specification networks and possible ways of how to obtain them in Section~\ref{sec:specification-networks}.
In Section~\ref{sec:neuro-symbolic-properties}, we then propose a fragment of quantifier-free first-order logic to formalize correctness properties involving specification networks.
We call this type of properties \emph{neuro-symbolic} and the resulting assertion language \nsal.

Once we have defined our new assertion language, Section~\ref{sec:reduction-to-deductive-verification} demonstrates how checking \nsal properties can be reduced to deductive verification of neural networks.
This reduction allows utilizing any existing deductive verifier (e.g., Planet~\cite{DBLP:conf/atva/Ehlers17}, Reluplex~\cite{DBLP:conf/cav/KatzBDJK17}, or Marabou~\cite{DBLP:conf/cav/KatzHIJLLSTWZDK19}), making our neuro-symbolic verification framework easily accessible to researchers and practitioners alike.
It is worth pointing out that other neural network verification techniques can also be lifted to neuro-symbolic verification, but we leave this research direction for future work.

%---------- Specification Networks ----------
\subsection{Specification Networks}
\label{sec:specification-networks}

Generally speaking, a specification network is a highly specialized deep neural network trained for a specific task (e.g., perception, anomaly detection, recognizing the underlying data distribution, etc.).
We use one (or multiple) of such networks as proxy objects to capture correctness properties.
Their precise architecture does not matter at this point, but might influence the choice of which verification engine to use.

Let us illustrate the concept of specification networks using the autonomous vehicle example from Section~\ref{sec:background}.
For the sake of simplicity, let us assume that we are given
\begin{itemize}
	\item a deep neural network $f$ that takes pictures $\vec{x}$ from the front camera as input and outputs the steering commands ``left'', ``right'', ``accelerate'', and ``decelerate''; and
	\item a property $P$ stating ``$f$ has to issue a deceleration command as soon as a stop sign appears in the front camera''.
\end{itemize}
Instead of trying to formalize all characteristics of stop signs in logic (i.e., their possible positions, shapes, colors, etc.), we now train a second deep neural network $g$ for the specific perception task of recognizing stop signs.
Assuming that $g$ is a binary-class network (outputting ``yes'' if it detects a stop sign in the image $\vec{x}$ and ``no'' otherwise), one can then express the property $P$ above straightforwardly as
\begin{align}
	\label{eq:stop-sign}
	\text{if $g(\vec{x}) = \text{``yes''}$, then $f(\vec{x}) = \text{``decelerate''}$.}
\end{align}
Note that our original property  $P$ now amounts to a simple constraint over the inputs and outputs of the networks $f$ and $g$.

An essential requirement of our framework is the availability of adequate specification networks.
We here sketch three conceivable ways of how to obtain them:
\begin{enumerate}
	\item
	The perhaps simplest way of obtaining specification networks is to train them explicitly.
	To avoid systematic errors, it is crucial to train a specification network on a dataset that is different from the one used to train the network under verification.
	Preferably, one should additionally use a different architecture and hyperparameters.
	\item
	Similar to standard datasets such as MNIST~\cite{lecun2010mnist}, researchers and interested companies might create public repositories for specification networks.
	To boot-strap such efforts, we have made the specification networks used in our experimental evaluation (see Section~\ref{sec:evaluation}) available at \url{https://github.com/LebronX/Neuro-Symbolic-Verification}.
	\item
	Finally, regulatory bodies might provide specification networks as references for future AI-enabled systems.
	Such an approach can be used, for instance, to guarantee minimum standards for the correctness and reliability of neural networks in safety-critical applications.
	Similarly, notified bodies, such as Germany's TÜV\footnote{ \url{https://www.tuv.com/world/en/} or \url{https://www.tuvsud.com/en}}, might provide specification networks as part of their testing, certification, and advisory services.
\end{enumerate}

%---------- Neuro-Symbolic Properties ----------
\subsection{A Neuro-Symbolic Assertion Language}
\label{sec:neuro-symbolic-properties}

Inspired by neuro-symbolic reasoning~\cite{DBLP:journals/flap/GarcezGLSST19,DBLP:conf/ijcai/RaedtDMM20}, we now describe how to use specification networks to formalize correctness properties of neural networks.
Specifically, we introduce an assertion language, named Neuro-Symbolic Assertion Language, which is inspired by the Hoare logic used in software verification~\cite{DBLP:journals/cacm/Hoare69} and follows the notation introduced by Albarghouthi~\cite{albarghouthi-book}.
This language is a fragment of the quantifier-free first-order logic over the reals and allows formalizing complex correctness properties---involving multiple specification networks---in an interpretable and straightforward manner.

Throughout the remainder of this paper, we assume that we are given $k \in \mathbb N$ specification networks $g_1, \ldots, g_k$ with $g_i \colon \mathbb R^{m_i} \to \mathbb R^{n_i}$ for $i \in \{ 1, \ldots, k \}$.
Moreover, let us assume that we want to formalize a correctness property for a single deep neural network $f \colon \mathbb R^{m_0} \to \mathbb R^{n_0}$, which we call the \emph{network under verification~(NUV)}.
Note that the latter assumption is not a restriction of our framework, but it simplifies the following presentation.
Our framework can easily be extended to multiple networks under verification.

Let us now turn to the definition of our \emph{Neuro-Symbolic Assertion Language (\nsal)}.
Formally, \nsal is the quantifier-free fragment of first-order logic over the reals that contains all logic formulas of the form
\begin{gather*}
    \bigl \{ \pre(\vec{x}_1, \ldots, \vec{x}_\ell) \bigr\} \\
    \mathrel{\vec{y}_1 \gets h_1(\vec{x}_1) \land \cdots \land \vec{y}_\ell \gets h_\ell(\vec{x}_\ell)} \\
    \bigl \{ \post(\vec{x}_1, \ldots, \vec{x}_\ell, \vec{y}_1, \ldots, \vec{y}_\ell) \bigr\},
\end{gather*}
where 
\begin{itemize}
	\item $h_i \in \{ f, g_1, \ldots, g_k \}$ for $i \in \{ 1, \ldots, \ell \}$ are function symbols representing the given neural networks, one of which is the NUV $f$;
	\item $\vec{x}_1, \ldots, \vec{x}_\ell$ are vectors of real variables representing the input values of the networks $h_i \in \{ f, g_1, \ldots, g_k \}$;
	\item $\vec{y}_1, \ldots, \vec{y}_\ell$ are vectors of real variables representing the output values of the networks $h_i \in \{ f, g_1, \ldots, g_k \}$;
	\item the expressions $\vec{y}_i \gets h_i(\vec{x}_i)$ store the result of the computation $h_i(\vec{x}_i) $ in the variable $\vec{y}_i$, where we assume that $\vec{x}_i$ and $\vec{y}_i$ match the dimensions of the input and output space of $h_i$, respectively;
	\item $\pre$ is a quantifier-free first-order formula over the free variables $\vec{x}_1, \ldots, \vec{x}_\ell$, called \emph{pre-condition}, expressing constraints on the inputs to the networks $f, g_1, \ldots, g_k$; and
	\item $\post$ is a quantifier-free first-order formula over the free variables $\vec{x}_1, \ldots, \vec{x}_\ell$ and $\vec{y}_1, \ldots, \vec{y}_\ell$, called \emph{post-condition}, expressing desired properties of $f$ while considering the computations of $g_1, \ldots, g_k$.
\end{itemize}
We call each such formula a \emph{neuro-symbolic property} to emphasize that correctness properties are no longer restricted to simple first-order constraints on the inputs and outputs of the network under verification but can depend on other networks.

The intuitive meaning of a neuro-symbolic property is that if the inputs $\vec{x}_1, \ldots, \vec{x}_\ell$ satisfy $\pre$ and the output of the networks on these inputs is $\vec{y}_1, \ldots, \vec{y}_\ell$, then $\post$ has to be satisfied as well.
Let us illustrate this definition with our example of Section~\ref{sec:specification-networks}.
In this example, we are given a NUV $f \colon \mathbb R^{m \times m} \to \bigl\{ \text{left}, \text{right}, \text{accelerate}, \text{decelerate} \bigr\}$ mapping $m \times m$ pixel images to steering commands and a single specification network $g \colon \mathbb R^{m \times m} \to \bigl\{ \text{yes}, \text{no} \bigr\}$ detecting stop signs.
Then, Property~\eqref{eq:stop-sign} can be formalized in \nsal as
\begin{gather*}
    \bigl\{ \vec{x}_1 = \vec{x}_2 \bigr\} \\
    \mathrel{ y_1 \gets f(\vec{x}_1) \land y_2 \gets g(\vec{x}_2)} \\
    \bigl\{ y_2 = \text{yes} \rightarrow y_1 = \text{decelerate} \bigr\}.
\end{gather*}

This neuro-symbolic property is a prototypical example of how our approach mitigates the first shortcoming of classical neural network verification discussed in Section~\ref{sec:background}.
To address the second and third shortcomings, we can train an autoencoder $g \colon \mathbb R^m \to \mathbb R^m$ to capture the distribution underlying the training data.
To restrict the verification of a network $f \colon \mathbb R^m \to \mathbb R^n$ to the underlying data distribution, we can use the neuro-symbolic property
\[ \bigl\{ \mathit{true} \bigr\} \mathrel{\vec{y}_1 \gets f(\vec{x}) \land \vec{y}_2 \gets g(\vec{x})} \{ \mathit{d}(\vec{x}, \vec{y}_2) \leq \varepsilon \rightarrow P(\vec{x}, \vec{y}_1) \}. \]
Here, $P$ is the original property we want to verify, and the condition $\mathit{d}(\vec{x}, \vec{y}_2) \leq \varepsilon$ for some $\varepsilon \geq 0$ follows the usual idea that a large reconstruction error (i.e., $\mathit{d}(\vec{x}, \vec{y}_2) > \varepsilon$) indicates out-of-distribution data~\cite{DBLP:conf/pricai/SakuradaY14}.
As a byproduct, we obtain that any counterexample to this new property violates the original property $P$ and originates from the underlying data distribution (as captured by the autoencoder $g$).

It is not hard to verify that ``simple'' properties, such as adversarial robustness and fairness, can easily be expressed as neuro-symbolic properties as well.
For instance, adversarial robustness can be formalized in \nsal as
\[ \bigl\{ \mathit{d}(\vec{x}^\star, \vec{x}) \leq \varepsilon \bigr\} \mathrel{\vec{y}^\star \gets f(\vec{x}^\star) \land \vec{y} \gets f(\vec{x})} \bigl\{ \vec{y}^\star = \vec{y} \bigr\} \]
where $\vec{x}^\star \in \mathbb R^m$ is a fixed input, $\epsilon \geq 0$, and assuming that the distance function $\mathit{d}$ can be expressed in the quantifier-free fragment of first-order logic over the reals.
Note that we allow individual networks to appear multiple times in a neuro-symbolic property.

Given a neuro-symbolic property $\{ \pre \} \mathrel{\assign} \{ \post \}$ with $\assign \coloneqq \bigwedge_{i = 1}^\ell \vec{y}_i \gets h_i(\vec{x}_i)$, the overall goal is to check whether the logic formula
\[ \psi ~\coloneqq~ \bigl( \pre \land \assign \bigr) \to \post \]
is valid (i.e., whether $\forall \vec{x}, \vec{y} \colon \psi$ is a tautology).
In analogy to software verification, we call this task the \emph{neuro-symbolic verification problem} and the formula $\psi$ a \emph{neuro-symbolic verification condition}.
The next section demonstrates how this verification problem can be reduced to deductive verification.

% !Tex root=main.tex

%---------- Reduction to Deductive Verification ----------
\section{Reduction to Deductive Verification}
\label{sec:reduction-to-deductive-verification}
In this section, we show how to translate a neuro-symbolic property, including the network under verification and the specification networks, into a (neuro-symbolic) verification condition, whose validity we can then check using an existing constraint solver.
This process is inspired by modern software verification, where it is called \emph{deductive verification}.
However, before we can describe our deductive verification approach in detail, we need to set up additional notation.

% Background and Notation
\subsection{Background and Notation}
For the purpose of deductive verification, we view a deep neural network $f \colon \mathbb R^m \to \mathbb R^n$ as an extended graph $G_f = (V, V_I, V_O, E, \alpha)$ where $V$ is a finite set of vertices (i.e., neurons), $V_I$ are the input neurons, $V_O$ are the output neurons (with $V_I \cap V_O = \emptyset)$, $E \subseteq V \times \mathbb R \times V$ is a weighted, directed edge relation, and $\alpha$ is a mapping that assigns an activation function (e.g., ReLU, sigmoid, etc.) to each neuron in $V \setminus V_I$.
Without loss of generality, we omit the network's biases since they can easily be included in the definition of the activation functions.
Moreover, we assume that the input neurons $V_I = \{ \nu_{\mathit{in}, 1}, \ldots, \nu_{\mathit{in}, m} \}$ and the output neurons $V_O = \{ \nu_{\mathit{out}, 1}, \ldots, \nu_{\mathit{out}, n} \}$ are implicitly ordered, reflecting the order of the inputs and outputs of the network.

%Before we can describe out translation in detail, we first need to introduce additional notation.
%For a given neuron $\nu \in V$, let $E(\nu) \coloneqq \{ \nu' \in V \mid \exists w \in \mathbb R \colon(v, w, v') \in E \}$ denote the set of successors of $\nu$ and $E^{-1}(\nu) \coloneqq \{ \nu' \in V \mid \exists w \in \mathbb R \colon(v', w, v) \in E \}$ the set of predecessors of $\nu$.
%This notation allows us to define the set of input neurons straightforwardly by $I(G_f) = \{ \nu \in V \mid E^{-1}(\nu) = \emptyset \}$ (i.e., all neurons that have no predecessor).
%Similarly, we can define the set of output neurons by $O(G_f) = \{ \nu \in V \mid E(\nu) = \emptyset \}$ (i.e., all neurons without a successor).
%For the sake of simplicity, we assume that neurons are implicitly ordered so that 

We also need to introduce elementary background on constraint solving.
To this end, let $\mathcal X$ be a set of real variables and $\varphi$ a quantifier-free first-order formula over $\mathcal X$.
Moreover, let $\mathcal I \colon \mathcal X \to \mathbb R$ be a mapping that assigns a real value to each variable in $\mathcal X$, called an \emph{interpretation}.
We define satisfaction as usual (see any textbook on first-order logic for details, for instance, the one by Huth and Ryan~\shortcite{DBLP:books/daglib/0000773}) and write $\mathcal I \models \varphi$ if the interpretation $\mathcal I$ \emph{satisfies} the formula $\varphi$ (i.e., the interpretation of the variables in $\varphi$ make the formula true).
Although checking the satisfiability of a first-order formula is undecidable in general, a host of effective constraint solvers for specific fragments of first-order logic exist. The verification conditions we generate in the following fall in such a fragment.

% Deductive Verification
\subsection{Deductive Verification of Neuro-Symbolic Properties}
To simplify the following presentation, let us assume for now that a given neuro-symbolic property $P$ involves a single network under verification $f \colon \mathbb R^m \to \mathbb R^n$ and no specification network (we explain shortly how this restriction can be lifted).
More precisely, let the property be given by
\[ P ~\coloneqq~ \bigl\{ \pre(\vec{x}) \bigr\} \mathrel{\assign(\vec{x}, \vec{y})} \bigl\{ \post(\vec{x}, \vec{y}) \bigr\}, \]
where $\vec{x} = (x_1, \ldots, x_m)$, $\vec{y} = (y_1, \ldots, y_n)$, and $\assign(\vec{x}, \vec{y}) \coloneqq \vec{y} \gets f(\vec{x})$.
Moreover, let $G_f = (V, V_I, V_O, E, \alpha)$ be the graph representation of $f$.

The key idea of our translation of $P$ into a neuro-symbolic verification condition is to substitute the appearance of the symbol $f$ in $P$ with a logic formula capturing the semantics of the deep neural network.
To this end, we assign to each neuron $\nu \in V$ a real variable $X_\nu$ that tracks the output of $\nu$ for a given input to the network.
For input neurons $\nu \in V_I$, the variables $X_\nu$ are simply used to store the input values of the neural network.
For all other neurons $\nu \in V \setminus V_I$, the variables $X_\nu$ are used to compute the output of $\nu$ given the variables of the neurons in the preceding layer.
This computation can be ``executed'' by constraining the variable $X_\nu$ using the formula
\[ \varphi_\nu ~\coloneqq~ X_\nu = \alpha(\nu) \Biggl( \sum_{(v', w, v) \in E} w \cdot X_{v'} \Biggr), \]
which first computes the weighted sum of the inputs to $\nu$ and then applies $\nu$'s activation function $\alpha(\nu)$.
In the case of a ReLU activation function, for instance, the formula $\varphi_\nu$ can be implemented straightforwardly as
\begin{multline}
	\label{eq:ReLU-constraint}
	\biggl( Y_\nu = \sum_{(v', w, v) \in E} w \cdot X_{v'} \biggr) \land {} \\
	\biggl( \bigl(Y_\nu \leq 0 \rightarrow X_\nu = 0 \bigr) \land \bigl(Y_\nu > 0 \rightarrow X_\nu = Y_\nu \bigr) \Biggr),
\end{multline}
where $Y_\nu$ is an unused, auxiliary variable.
Note that $\varphi_\nu$ falls into the fragment of Linear Real Arithmetic~(LRA) in this specific case.

To capture the semantics of the entire network $f$, we can simply take the conjunction
\[ \varphi_f ~\coloneqq~ \bigwedge_{\nu \in V \setminus V_I} \varphi_\nu \]
of all neuron constraints defined above.
A straightforward induction over the layers of $f$ then shows that this formula indeed simulates the computation of $f$ on any input, as formalized in the lemma below.
Note that $\varphi_f$ ranges over the variables $X_\nu$ for all neurons $\nu \in V$, even the input neurons.

\begin{lemma} \label{lem:verification-condition}
Let $f \colon \mathbb R^m \to \mathbb R^n$ be a deep neural network with graph representation $G_f = (V, V_I, V_O, E, \alpha)$, $\vec{x} = (x_1, \ldots, x_m) \in \mathbb R^m$ an input to $f$, and $\mathit{out}_\nu \in \mathbb R$ the output of neuron $\nu \in V$ when $f$ processes the input $\vec{x}$.
Moreover, let $\varphi_f$ be as defined above and $\mathcal I$ an interpretation with $\mathcal I(X_{\nu_{\mathit{in}, i}}) = x_i$ for $i \in \{ 1, \ldots, m \}$.
Then, $\mathcal I \models \varphi_f$ if and only if $\mathcal I(X_\nu) = \mathit{out}_\nu$ for each neuron $\nu \in V$.
In particular, $f(\vec{x}) = \bigl(I(X_{\nu_{\mathit{out}, 1}}), \ldots, I(X_{\nu_{\mathit{out}, n}}) \bigr)$ holds (i.e., the output of $f$ on $\vec{x}$ can be obtained from the variables $X_{\nu_{\mathit{out}, 1}}, \ldots, X_{\nu_{\mathit{out}, n}}$).
\end{lemma}

Given the variables $X_\nu$ for $\nu \in V$, the construction of the neuro-symbolic verification condition is now straightforward.
First, we replace the formula $\assign$ with $\varphi_f$.
Second, we substitute each occurrence of the variables $\vec{x} = (x_1, \ldots, x_m)$ and $\vec{y} = (y_1, \ldots, y_n)$ in the formulas $\pre$ and $\post$ by $\vec{X}_{V_I} = (X_{\nu_{\mathit{in}, 1}}, \ldots, X_{\nu_{\mathit{in}, m}})$ and $\vec{X}_{V_O} = (X_{\nu_{\mathit{out}, 1}}, \ldots, X_{\nu_{\mathit{out}, n}})$, respectively.
This process then results in the verification condition
\[ \psi ~\coloneqq~ \bigl( \pre[\vec{x}\,/\,\vec{X}_{V_I}] \land \varphi_f \bigr) \rightarrow \post[\vec{x}\,/\,\vec{X}_{V_I}, \vec{y}\,/\,\vec{X}_{V_O}], \]
where we use $\varphi[\vec{z}_1\,/\,\vec{z_2}]$ to denote the formula resulting from the substitution of the vector of variables $z_1$ by $z_2$ in $\varphi$.

In order to determine whether the network $f$ satisfies the neuro-symbolic property $P$, we now have to check whether $\psi$ is valid (i.e., satisfied by all possible values of the free variables).
Typically, this is done by checking whether the negation $\lnot \psi$ is satisfiable.
If $\psi$ is satisfiable (i.e., the property does not hold), a satisfying assignment $\mathcal I$ of $\psi$ can be used to derive inputs to $f$ (from $X_{\nu_{\mathit{in}, 1}}, \ldots, X_{\nu_{\mathit{in}, m}}$) that violate the property $P$.
The correctness of this approach follows from Lemma~\ref{lem:verification-condition} and is summarized in the theorem below.

\begin{theorem} \label{thm:deductive-verification}
Let $f$ be a deep neural network, $P$ a neuro-symbolic property, and $\psi$ the neuro-symbolic verification condition as constructed above.
Then, $f$ satisfies $P$ if and only if $\lnot \psi$ is satisfiable (i.e., $\psi$ is valid).
Additionally, if $\mathcal I \models \lnot \psi$, then $\mathcal I(X_{\nu_{\mathit{in}, 1}}), \ldots, \mathcal I(X_{\nu_{\mathit{in}, m}})$ are inputs to $f$ that violate $P$.
\end{theorem}

The approach described in this section can easily be generalized to neuro-symbolic properties that contain multiple specification networks $g_1, \ldots, g_\ell$ (and even multiple networks under verification).
In this case, the formula $\assign$ needs to be replaced by the conjunction $\varphi_f \land \varphi_{g_1} \land \cdots \land \varphi_{g_\ell}$ where $\varphi_f$ and $\varphi_{g_1}, \ldots, \varphi_{g_\ell}$ are constructed as described above.
Moreover, the variables $\vec{x}_1, \ldots, \vec{x}_\ell$ and $\vec{y}_1, \ldots, \vec{y}_\ell$ in the formulas $\pre$ and $\post$ now have to be replaced by their corresponding counterparts $\vec{X}_{V_I^1}, \ldots, \vec{X}_{V_I^\ell}$ and $\vec{X}_{V_O^1}, \ldots, \vec{X}_{V_O^\ell}$, respectively.
It is not hard to verify that a generalized version of Theorem~\ref{thm:deductive-verification} holds as well.

%---------- Modifying existing verification infrastructure ----------
\subsection{Building Neuro-Symbolic Verifiers on Top of Existing Verification Infrastructure}
\label{sec:nsv}

Let us now describe how to build a neuro-symbolic verifier on top of the existing verification infrastructure for deep neural network verification.
After translating the given neuro-symbolic property and all deep neural networks into a neuro-symbolic verification condition $\psi$, the remaining task is to check the satisfiability of $\lnot \psi$ and extract a counterexample if the verification fails (i.e., $\lnot \psi$ is satisfiable).
Since our neuro-symbolic verification conditions fall into a decidable fragment of real arithmetic, one can simply apply off-the-shelf Satisfiability Modulo Theories solvers (e.g., dReal~\cite{DBLP:conf/cade/GaoAC12}, CVC4~\cite{DBLP:conf/cav/BarrettCDHJKRT11}, or Z3~\cite{DBLP:conf/tacas/MouraB08}).
On the level of logic, it is irrelevant whether a verification condition involves one or multiple neural networks and which of them are specification networks.
The only important property is that the resulting verification condition falls into a fragment of first-order logic that the constraint solver can handle. 

In addition to the ``general-purpose'' Satisfiability Modulo Theories solvers mentioned above, a range of specialized constraint solvers exists for the classical verification of deep neural networks with fully-connected layers or ReLU activation functions (e.g., Planet~\cite{DBLP:conf/atva/Ehlers17}, Reluplex~\cite{DBLP:conf/cav/KatzBDJK17}, and Marabou~\cite{DBLP:conf/cav/KatzHIJLLSTWZDK19}).
As long as the NUV and the specification networks are of this form, all three solvers can also be used for neuro-symbolic verification because Constraint~\eqref{eq:ReLU-constraint} appears only non-negated in $\lnot \psi \coloneqq \pre \land \assign \land \lnot \post$.
Consequently, the resulting verification conditions fall into a logical fragment that all three solvers can handle.

To demonstrate the ease of using existing infrastructure for neuro-symbolic verification, we have built a deductive verifier on top of the popular Marabou framework~\cite{DBLP:conf/cav/KatzHIJLLSTWZDK19}, called \emph{Neuro-Symbolic Verifier (\verifier)}.
Since the standard version of Marabou does not support \mbox{\nsal} properties---or even classical verification queries with multiple networks---, we have modified it as follows:
\begin{itemize}
    \item
    We have extended Marabou's input language to support multiple deep neural networks and correctness properties expressed in \nsal.
    \item 
    We have added a lightweight bookkeeping mechanism to track which of the variables\slash constraints corresponds to which neural network.
    \item
    We have extended Marabou's reporting facilities to extract counterexamples that consist of inputs to multiple networks or relate multiple networks (using the bookkeeping mechanism described above).
\end{itemize}
These modifications did not require any substantial changes to the core of Marabou, showing that our neuro-symbolic verification framework can effortlessly be adopted in practice.

%---------- Evaluation ----------
% !Tex root=main.tex

\section{Empirical Evaluation}
\label{sec:evaluation}

In this section, we demonstrate that \verifier is effective in verifying a variety of neuro-symbolic properties.
However, it is paramount to stress that we are not interested in the absolute performance of \verifier, how well it scales to huge networks, or how it compares to other verification techniques on non-neuro-symbolic properties.
Instead, our goals are twofold:
\begin{enumerate*}[label={(\arabic*)}]
    \item we demonstrate that our neuro-symbolic approach can be implemented on top of existing verification infrastructure for deep neural networks and is effective in verifying neuro-symbolic properties; and
    \item we showcase that our neuro-symbolic framework can find more informative counterexamples than a purely deductive verification approach in case the verification fails.
\end{enumerate*}
Note that the former makes it possible to leverage future progress in neural network verification to our neuro-symbolic setting, while the latter greatly improves the debugging of learning systems. 

In our experimental evaluation, we have considered two widely used datasets:
\begin{enumerate}
	\item
	The MNIST dataset~\cite{lecun2010mnist}, containing 60,000 training images and 10.000 test images of ten hand-written digits.
	\item
	The German Traffic Sign Recognition Benchmark (\mbox{GTSRB})~\cite{DBLP:conf/ijcnn/StallkampSSI11}, containing 39,209 training images and 12,630 test images with 43 types of German traffic signs.
	To not repeat similar experiments too often, we have restricted ourselves to the first ten (of the 43) classes.
\end{enumerate}
For both datasets, the task is to predict the correct class of an input image (i.e., which sign or which digit).

The remainder is structured along the two goals laid out at the beginning of this section.
We first demonstrate the effectiveness of \verifier in verifying neuro-symbolic properties and then show that \verifier can generate more informative counterexamples to failed verification attempts.
The code of \verifier and all experimental data can be found online at \url{https://github.com/LebronX/Neuro-Symbolic-Verification}.

%---------- Effectiveness of Verifying Neuro-Symbolic Properties ----------
\subsection{Effectiveness of Verifying Neuro-Symbolic Properties}
For both the GTSRB and the MNIST datasets, we considered the following three prototypical neuro-symbolic properties.
By convention, we use $f$ to denote the network under verification (NUV) and $g$ to denote a specification network.
Moreover, we use the $L_\infty$-norm as distance function $\mathit{d}$.
\begin{itemize}
    \item[$P_1$:]
    ``If the input image $\vec{x}$ is of class $c$, then the NUV outputs~$c$'', expressed in \nsal as
    \begin{gather*}
        \bigl\{ \mathit{true} \bigr\} \\
        \mathrel{\vec{y}_1 \gets f(\vec{x}) \land y_2 \gets g(\vec{x})} \\
        \bigl\{ y_2 = 1 \rightarrow \argmax{(\vec{y}_1)} = c \bigr\}.
    \end{gather*}
    Here, the NUV $f$ is a multi-class deep neural network mapping images to their class (i.e., one of the 43 traffic signs or one of the ten digits), while the specification network $g$ is a deep neural network specifically trained to detect the specific class $c$ (outputting $0$ or $1$).%
    \footnote{
	    This property is a simplified version of Property~\eqref{eq:stop-sign} on Page~\pageref{eq:stop-sign}.
	    Note that we here use a specification network to ``decide'' the class of an image instead of the actual label.
    }
    \item[$P_2$:]
    ``If the input $\vec{x}$ follows the distribution of the underlying data, then the NUV classifies the input correctly with high confidence'', expressed in \nsal as
    \begin{gather*}
        \bigl\{ \mathit{true} \bigr\} \\
        \mathrel{\vec{y}_1 \gets f(\vec{x}) \land \vec{y}_2 \gets g(\vec{x})} \\
        \bigl\{ (\mathit{d}(\vec{y}_2 - \vec{x}) \leq \varepsilon \land \argmax{(\vec{y}_1)} = c) \rightarrow \mathit{conf} > \delta \bigr\},
    \end{gather*}
    where $\varepsilon, \delta \geq 0$ and $\mathit{conf} \coloneqq \nicefrac{(|\vec{y}_1| \cdot y_i - \sum_{j \neq i} y_j)}{|\vec{y}_1|}$ is the confidence of the NUV that the input $\vec{x}$ is of class $c$.
    Here, the NUV $f$ is a multi-class deep neural network mapping images to their class (i.e., one of the 43 traffic signs or one of the ten digits), while the specification network $g$ is an autoencoder used to detect out-distribution data~\cite{DBLP:conf/pricai/SakuradaY14} (i.e., if $\mathit{d}(\vec{y}_2 - \vec{x}) > \varepsilon$).
    \item[$P_3$:]
    ``Two deep neural networks (of different architecture) compute the same function up to a maximum error of $\varepsilon$'', expressed in \nsal as
    \begin{gather*}
        \bigl\{ \mathit{true} \bigr\}
        \mathrel{\vec{y}_1 \gets f(\vec{x}) \land \vec{y}_2 \gets g(\vec{x})}
        \bigl\{ \mathit{d}(\vec{y}_1 - \vec{y}_2) \leq \varepsilon \bigr\},
    \end{gather*}
    where $\varepsilon \geq 0$.
    Here, the NUV $f$ and the specification network $g$ have the same dimensions of the input and output space but potentially distinct architectures.
\end{itemize}

For each benchmark suite, each class (remember that we have only considered ten classes of GTSRB), and each of the three properties, we have trained one NUV and one specification network with the architectures shown in Table~\ref{tab:architecture} (all using ReLU activation functions).
We have resized the MNIST images for property $P_2$ to $14 \times 14$ and the \mbox{GTSRB} images to $16 \times 16$ for all properties to keep the verification tractable.
For property $P_2$, we have chosen $0.05 \leq \varepsilon \leq 0.14$ with step size $0.01$ and $1 \leq \delta \leq 20$ with step size $1$.
For property $P_3$, we have chosen $0.05 \leq \varepsilon \leq 0.14$ with step size $0.01$.

\begin{table}[!t]
    \centering
    \small
    \begin{tabular}{l*{3}{r}*{3}{r}}
        \toprule
        Property and & \multicolumn{3}{c}{NUV} &  \multicolumn{3}{c}{Spec.\ network} \\ \cmidrule(lr){2-4} \cmidrule(lr){5-7}
        benchmarks & in & out & hid. & in & out & hid. \\ \midrule
        $P_1$-MNIST & $784$ & $10$ & $3 \cdot 10$ & $784$ & $2$ & $3 \cdot 10$ \\
        $P_2$-MNIST & $196$ & $10$ & $1 \cdot 10$ & $196$ & $196$ & $1 \cdot 10$ \\
        $P_3$-MNIST & $784$ & $10$ & $3 \cdot 20$ & $784$ & $10$ & $3 \cdot 20$ \\ \midrule
        $P_1$-GTSRB & $256$ & $43$ & $3 \cdot 12$ & $256$ & $2$ & $3 \cdot 5$ \\
        $P_2$-GTSRB & $256$ & $43$ & $1 \cdot 10$ & $256$ & $256$ & $1 \cdot 10$ \\
        $P_3$-GTSRB & $256$ & $43$ & $3 \cdot 35$ & $256$ & $43$ & $3 \cdot 35$ \\
        \bottomrule
    \end{tabular}
    \caption{Architectures used in our experimental evaluation. Each line lists the number of neurons in the input layer (``in''), output layer (``out''), and hidden layers (``hid.''), respectively.}
    \label{tab:architecture}
\end{table}

To avoid statistical anomalies, we have repeated all experiments five times with different neural networks (trained using different parameters) and report the average results.
This way, we obtained $5 \cdot 2 \cdot (10 + 200 + 10) = 2,200$ verification tasks in total.
We have conducted our evaluation on an Intel Core i5-5350U CPU (1.80\,GHz) with 8\,GB RAM running MacOS Catalina 10.15.7 with a timeout of $1,800\,s$ per benchmark.

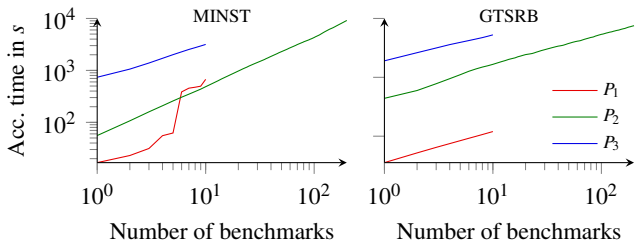
\begin{figure}[!th]
    \centering
    \begin{tikzpicture}
        \begin{groupplot}
        [
            group style={
                group size=2 by 1,
                ylabels at=edge left,
                yticklabels at=edge left,
                horizontal sep=0.5cm
            },
            /tikz/font=\footnotesize,
            width=49mm,
            height=35mm,
            xlabel={Number of benchmarks},
            ylabel={Acc.\ time in $s$},
            axis lines=left,
            tick align=outside,
            title style={font=\scriptsize, at={(axis description cs:0.5, .8), anchor=north, inner sep=0pt}},
            legend style={draw=none, fill=none, font=\scriptsize, at={(1, 0)}, anchor=south east},
            cycle list name=mylist,
            axis on top,
            xmode=log,
            ymode=log,
            xmax=200,
            ymax=10000,
        ]

            \nextgroupplot[title={MINST}, ]
                \addplot table [x expr=\coordindex+1, y={accumulated time}, col sep=comma] {csv/mnist-P1.csv};

                \addplot table [x expr=\coordindex+1, y={accumulated time}, col sep=comma] {csv/mnist-P2.csv};
            
                \addplot table [x expr=\coordindex+1, y={accumulated time}, col sep=comma] {csv/mnist-P3.csv};
            
            \nextgroupplot[title={GTSRB}]
                \addplot table [x expr=\coordindex+1, y={accumulated time}, col sep=comma] {csv/gtsrb-P1.csv};

                \addplot table [x expr=\coordindex+1, y={accumulated time}, col sep=comma] {csv/gtsrb-P2.csv};
            
                \addplot table [x expr=\coordindex+1, y={accumulated time}, col sep=comma] {csv/gtsrb-P3.csv};
            
            \legend{$P_1$, $P_2$, $P_3$};
            
        \end{groupplot}
    \end{tikzpicture}
    \caption{Accumulated average runtimes for the experiments on the MNIST dataset (left) and the GTSRB dataset (right)}
    \label{fig:accumulative-runtimes}
\end{figure}

Figure~\ref{fig:accumulative-runtimes} depicts the results of our experiments in terms of the accumulated average runtimes.
On the MNIST benchmark suite, \verifier timed out on one benchmark (for property $P_1$) and terminated on all others.
It found a counterexample in all cases (i.e., none of the NUVs satisfied the properties).
On the GTSRB suite, \verifier always terminated.
It proved that all NUVs satisfied property $P_1$, while  finding counterexamples for all benchmarks of properties $P_2$ and $P_3$.
Note that the single timeout on the MINST benchmark suite causes the steep increase in the graph of property $P_1$ on the left of Figure~\ref{fig:accumulative-runtimes}.
Moreover, note that we have not taken any measures during the training process to ensure that our NUVs satisfy any of the properties, which explains the large number of counterexamples.
We believe that the relatively low resolution of the GTSRB images causes the NUVs to satisfy property $P_1$.

\pgfplotscreateplotcyclelist{mylist}{
    {red!90!black, mark=none},
    {green!50!black, mark=none},
    {blue!90!black, mark=none}
}

In total, our experiments show that \verifier is a versatile tool, effective at verifying a diverse set of neuro-symbolic properties.
The fact that it was built on top of existing verification infrastructure further shows that our neuro-symbolic framework is easy to adopt in practice, making it accessible to researchers and practitioners alike.

%---------- Quality of Counterexamples ----------
\subsection{Quality of Counterexamples}
To assess the quality of the counterexamples generated by \verifier, we have modified Property $P_2$ to exclude the requirement that the data must come from the underlying distribution.
The resulting property $P'_2$, expressed in \nsal, is
\begin{gather*}
    \bigl\{ \mathit{true} \bigr\}
    \mathrel{\vec{y}_1 \gets f(\vec{x}) }
    \bigl\{ \argmax{(\vec{y}_1)} = c \rightarrow \mathit{conf} > \delta \bigr\}.
\end{gather*}
Note that $P'_2$ involves only one network and represents a typical global property arising in classical neural network verification.

\begin{figure}[!t]
    \centering
    \includegraphics[width=.2\linewidth]{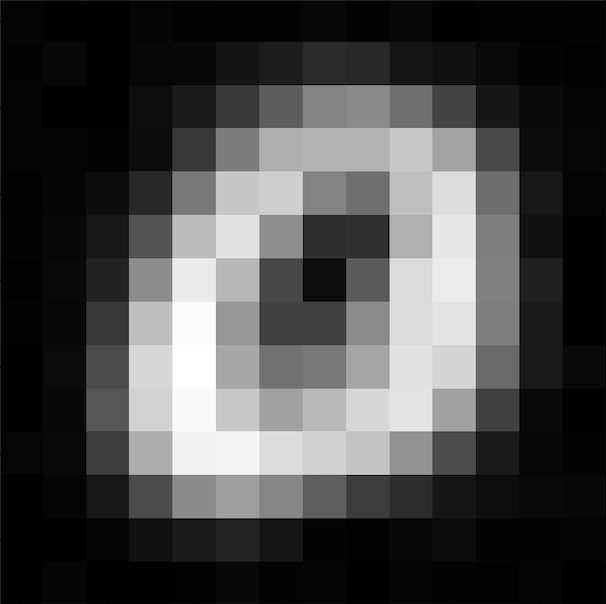}
    \includegraphics[width=.2\linewidth]{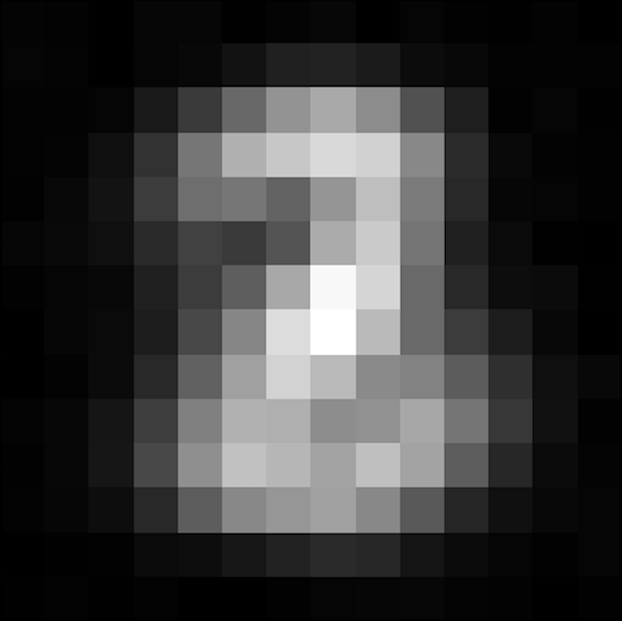}
    \hskip .1\linewidth
    \includegraphics[width=.2\linewidth]{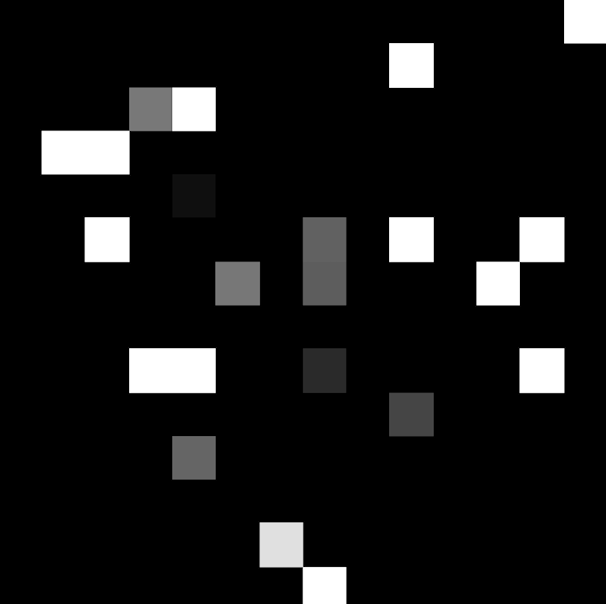}
    \includegraphics[width=.2\linewidth]{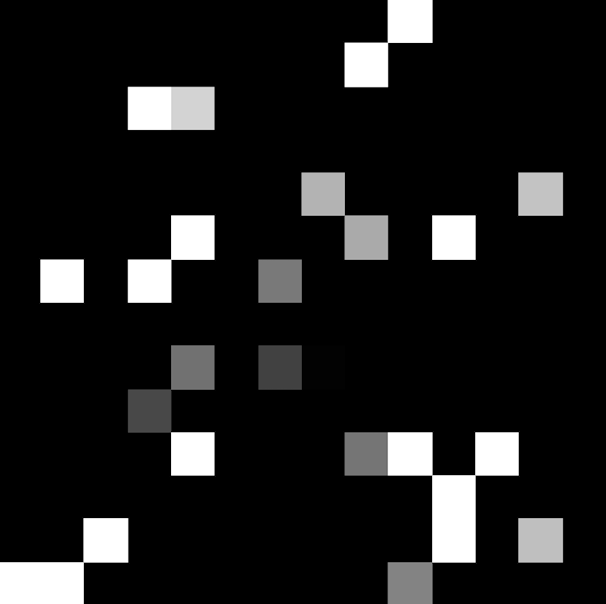}
    
    \caption{Counterexamples to the neuro-symbolic property $P_2$ (left) and the corresponding non-neuro-symbolic property $P'_2$ (right)}
    \label{fig:counterexamples}
\end{figure}

We have verified Property~$P'_2$ on the deep neural networks from the MNIST dataset using the original Marabou framework. 
Selected results are shown on the right-hand-side of Figure~\ref{fig:counterexamples}.
Since Property~$P'_2$ is global, the verification has to consider all possible inputs.
As Figure~\ref{fig:counterexamples} demonstrates, counterexamples to such properties are often random noise and arguably of little value.
In fact, we could not identify a single counterexample that looked close to the original dataset.

By contrast, the left-hand-side of Figure~\ref{fig:counterexamples} shows two counterexamples to the neuro-symbolic property $P_2$.
These counterexamples are substantially more meaningful and intuitive because they originate from the underlying distribution of the data (as captured by an autoencoder trained to reconstruct the data).
This demonstrates that neuro-symbolic verification produces meaningful counterexamples that can greatly simplify the development and debugging of learning systems.

%---------- Conclusion ----------
% !Tex root=main.tex

\section{Conclusion and Future Work}
Today's approaches to neural network verification are limited to ``simple'' properties that can be formalized by quantifier-free first-order constraints.
To mitigate this severe practical restriction, we have introduced the first neuro-symbolic framework for neural network verification, which allows expressing complex correctness properties through deep neural networks.
We have demonstrated that our framework can straightforwardly be implemented on top of existing verification infrastructure and provides more informative counterexamples than existing methods.
To the best of our knowledge, we are the first to propose a neuro-symbolic approach to formal verification.

The concept of neuro-symbolic verification can, in principle, also be applied to hardware and software verification (e.g., to express properties involving perception), and we believe that this is a promising direction of future work.
Another essential future task will be to develop novel verification algorithms (e.g., based on abstract interpretation) that exploit the neuro-symbolic nature of correctness properties and can reason about multiple, inter-depending deep neural networks.
To further improve scalability, we also intend to investigate neuro-symbolic approaches to (concolic) testing and symbolic execution.

%---------- Bibliography ----------
\bibliographystyle{named}
\bibliography{bib}

\end{document}